\documentclass{article}
\usepackage{spconf,amsmath,graphicx,hyperref}

\usepackage{graphicx}
\usepackage{multirow}
\usepackage{booktabs}

\usepackage{amsfonts}  
\usepackage{amssymb}   
\usepackage{xcolor}
\usepackage{color}
\usepackage{tcolorbox}

\newcommand{\method}{Fake-HR1}

\title{FAKE-HR1: RETHINKING REASONING OF VISION LANGUAGE MODEL FOR
SYNTHETIC IMAGE DETECTION}

\name{
\begin{tabular}{c}
    Changjiang Jiang$^{1,2}$ \qquad Xinkuan Sha$^{2}$ \qquad Fengchang Yu$^{1}$\sthanks{Corresponding author: yufc2002@whu.edu.cn} \qquad Jingjing Liu$^{2}$\sthanks{Corresponding author: jingjing.lqq@antgroup.com} \\
    Jian Liu$^{2}$\sthanks{Corresponding author: rex.lj@antgroup.com} \qquad Mingqi Fang$^{2}$ \qquad Chenfeng Zhang$^{2,3}$ \qquad Wei Lu$^1$
  \end{tabular}
}
  
\address{$^1$ Wuhan University\\
  $^2$AntGroup\\
  $^3$Zhejiang University
  }

\begin{document}
\ninept
\maketitle
\begin{abstract}
Recent studies have demonstrated that incorporating Chain-of-Thought (CoT) reasoning into the detection process can enhance a model's ability to detect synthetic images. However, excessively lengthy reasoning incurs substantial resource overhead, including token consumption and latency, which is particularly redundant when handling obviously generated forgeries. To address this issue, we propose Fake-HR1, a large-scale hybrid-reasoning model that, to the best of our knowledge, is the first to adaptively determine whether reasoning is necessary based on the characteristics of the generative detection task. To achieve this, we design a two-stage training framework: we first perform Hybrid Fine-Tuning (HFT) for cold-start initialization, followed by online reinforcement learning with Hybrid-Reasoning Grouped Policy Optimization (HGRPO) to implicitly learn when to select an appropriate reasoning mode. Experimental results show that Fake-HR1 adaptively performs reasoning across different types of queries, surpassing existing LLMs in both reasoning ability and generative detection performance, while significantly improving response efficiency.
\end{abstract}
\section{Introduction}
\label{sec:intro}

With the rapid development of diffusion models \cite{diffusionmodel}, AIGC technologies are increasingly integrating synthetic multimodal data into our daily lives. For instance, SORA \cite{sora_videoworldsimulators2024} can generate highly realistic videos, while Qwen-Image \cite{wu2025qwenimagetechnicalreport} is capable of understanding text and manipulating images. However, synthetic multimodal data also introduces significant risks, including potential misuse \cite{glff}. Such risks include the creation of forged watermark \cite{watemark} and deepfake \cite{du2025forensichubunifiedbenchmark,deepfake,ddvqa,su2025spase} through diffusion models, the synthesis of fraudulent faces for scams, and the contamination of internet training data \cite{ye2024loki}. Given the ease of generating synthetic content, the internet may in the future be inundated with AI-generated material, making the task of verifying the authenticity and reliability of multimodal data increasingly challenging.

To address these threats, the field of Synthetic Content Detection has recently garnered substantial attention \cite{zhu2025mesorch,yan2025sanitycheckaigeneratedimage,ma2025imdl}. Nevertheless, existing approaches are predominantly limited to binary classification, with restricted human interpretability of predictions \cite{wen2025spot}. The rapid emergence of large reasoning models (LRMs) \cite{jiang2025thinkneedlargehybridreasoning,fang2025thinklessllmlearnsthink} has sparked interest in their ability to detect synthetic multimodal reasoning data \cite{wen2025spot}. 

On one hand, VLMs can provide natural-language justifications for authenticity judgments, thereby enhancing interpretability. On the other hand, distinguishing between authentic and synthetic data requires multimodal perception, knowledge integration, and reasoning, making this task an ideal testbed for evaluating the capabilities of large multimodal models (LMMs). This study, therefore, seeks to improve the performance of LMMs on Synthetic Content Detection tasks.

Existing methods such as GenImage \cite{genimage} and Community Forensics \cite{Park_2025_CVPR} primarily focus on binary detection, providing insufficient evaluation of LRMs' reasoning capabilities in Synthetic Content Detection. While datasets such as FakeClue \cite{wen2025spot} and FakeBench \cite{li2024fakebench} is closer to our objectives, they require models to produce reasoning chains even for samples with obvious synthetic artifacts, leading to increased inference latency and computational cost. This raises a key question: What constitutes an appropriate learning objective for LRMs in Synthetic Content Detection? We argue that a sophisticated model should not be a rigid reasoning machine: (1) for synthetic content with clear artifacts, reasoning is unnecessary, and the model can directly output the final decision (real or fake); (2) for subtle and difficult-to-detect cases—particularly those generated by state-of-the-art diffusion models \cite{wu2025qwenimagetechnicalreport}—models should produce detailed reasoning chains to enhance interpretability and classification accuracy.

To fill this gap, we introduce \method, an LRM capable of adaptively deciding whether to generate reasoning chains based on the input image and query, thereby balancing efficiency and performance. Our contributions are follow:

\begin{itemize}
\item We propose a comprehensive hybrid-reasoning training framework that can be applied across multiple models, enabling the construction and training of hybrid reasoning models (HRMs) specifically tailored for AIGC Detection tasks.
\item We present \method, an HRM that adaptively determines whether reasoning is necessary, striking a balance between inference latency and detection accuracy.
\end{itemize}

\section{Method}

\begin{figure*}[t]
\begin{align}
\mathcal{J}_{GRPO}(\theta) = & \mathbb{E}_{\substack{(x,y) \sim \mathcal{D}_{\text {CoT}}, \{o_i\}_{i=1}^G \sim \pi_{\theta_{\text{old}}} (O \mid x)}} \biggl[ \frac{1}{G} \sum_{i=1}^G \min \biggl(\frac{\pi_\theta(o_i \mid x)}{\pi_{\theta_{\text{old}}}(o_i \mid x)} A_i,\\
& \operatorname{clip}\left(\frac{\pi_\theta(o_i \mid x)}{\pi_{\theta_{\text{old}}}(o_i \mid x)}, 1-\varepsilon, 1+\varepsilon\right) A_i\biggr) \notag  - \beta \mathbb{D}_{KL}(\pi_\theta \| \pi_{\mathrm{SFT}}) \biggr]
\end{align}
\end{figure*}

\subsection{Hybrid Fine-Tuning (HFT)}

The goal of HFT is to construct a model capable of mastering two distinct response modes—reasoning mode and non-reasoning mode. To this end, we propose a simple yet effective dual-mode data strategy, which systematically partitions existing datasets into reasoning and non-reasoning subsets, thereby avoiding the need for costly manual annotation.

\begin{figure}[htbp]
    \centering
    \includegraphics[width=0.95\columnwidth]{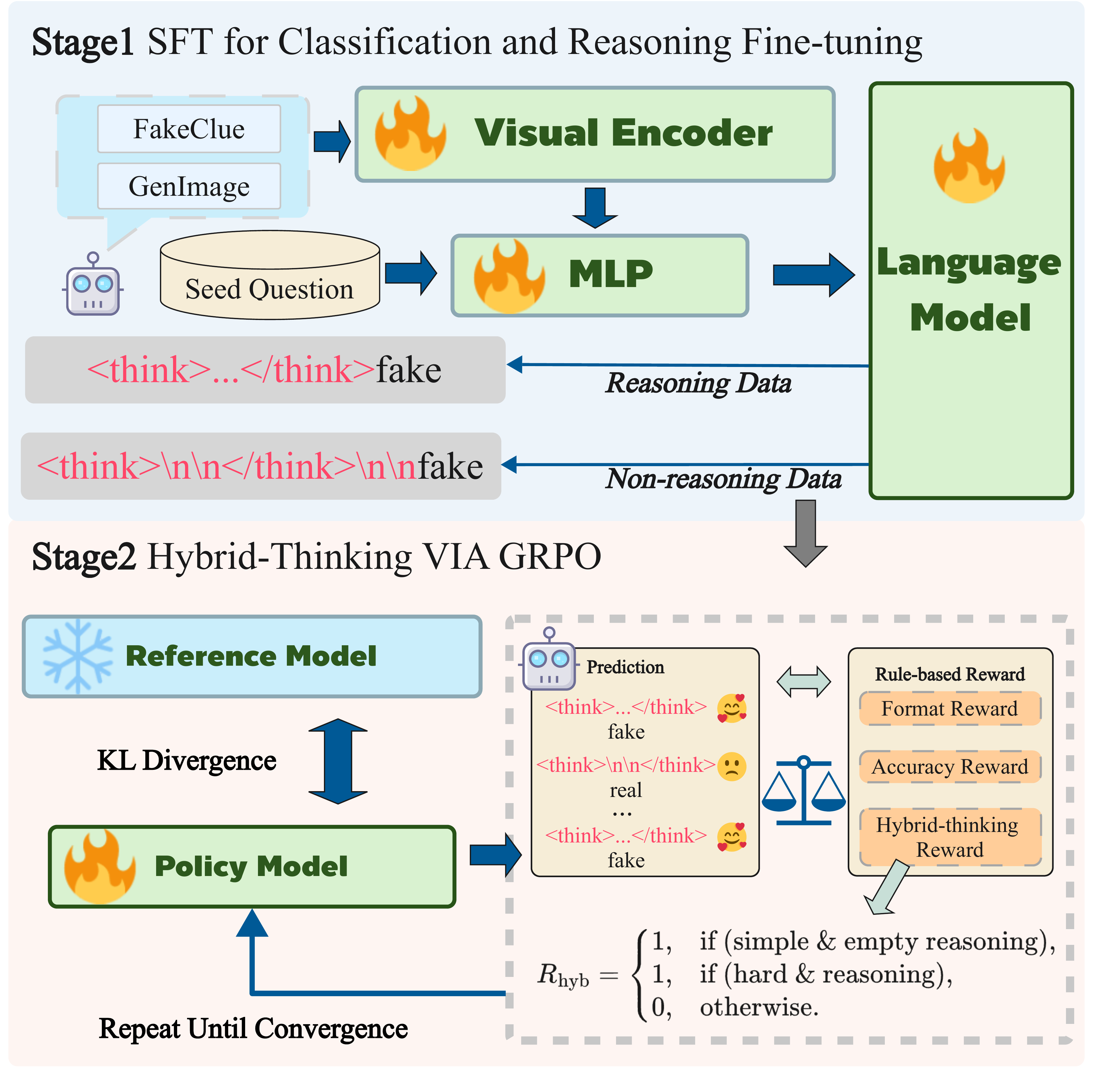}
    \caption{The training framework of \method.}
    \label{fig:method}
\end{figure}

As illustrated in Figure~\ref{fig:method}, our approach leverages two targeted heuristics depending on the query type:
(i) Data-oriented heuristic (low-cost acquisition of reasoning and non-reasoning data): existing datasets that already contain reasoning and non-reasoning responses are directly divided into two categories without requiring further annotation;
(ii) Query-oriented heuristic (objective query distinction): based on query complexity, we construct two seed question banks—reasoning-required questions and non-reasoning questions. These seed questions are used to systematically differentiate dual-mode data. Specifically, for datasets without reasoning types (e.g., GenImage \cite{genimage}), we randomly sample from the simple seed bank to populate non-reasoning queries, while for datasets containing reasoning-intensive samples (e.g., FakeClue), we apply the same strategy with the complex seed bank.

\subsection{Hybrid-Thinking VIA Group Reward Policy OPTIMIZATION}

HFT successfully endows $\method_{sft}$ with the dual abilities of reasoning and direct response. However, when $\method_{sft}$ operates under automatic reasoning settings, we observe performance degradation on certain datasets. This phenomenon, which we term “reasoning degeneration”, manifests as the model defaulting to non-reasoning responses even for complex queries requiring reasoning. Such failures in instruction-following suggest that while the model possesses the necessary skills, it lacks the judgment to deploy them appropriately.

Fortunately, Group Relative Policy Optimization (GRPO) \cite{shao2024deepseekmathpushinglimitsmathematical} provides a natural paradigm to address this issue. By optimizing policies based on outcome-driven rewards, GRPO enables the model to learn when and how to adopt the most effective reasoning strategy. However, directly applying GRPO introduces bias, as the model may develop a preference for a single reasoning mode. Existing hybrid-reasoning methods also face critical limitations: (1) reliance on overly complex reward models and handcrafted rules, and (2) rigid reasoning modes that are excessively data- and prompt-sensitive. To overcome these challenges, we extend GRPO to explicitly incentivize hybrid-thinking capacity.

\textbf{Group Relative Policy Optimization} As shown in Figure~\ref{fig:method}. Unlike traditional actor–critic methods, GRPO eliminates the critic network and instead estimates the baseline through group averaging, substantially reducing GPU memory consumption. For each input pair $(x, y)$, the policy $\pi_\theta$ samples a group of $G$ candidate responses ${o_i}_{i=1}^G$.

where $\varepsilon$ and $\beta$ are hyperparameters, and $\pi_{\mathrm{SFT}}$, $\pi_{\theta}$, and $\pi_{\theta_{\text{old}}}$ are the model after SFT, the optimized model and the old policy model.
The group-normalized advantage for the $i$-th response is:
\begin{equation}
A_i=\frac{r_i-\operatorname{mean}\left(\left\{r_1, r_2, \cdots, r_G\right\}\right)}{\operatorname{std}\left(\left\{r_1, r_2, \cdots, r_G\right\}\right)}
\end{equation}

\textbf{Reward Model} To guide RL training, we design a rule-based reward that integrates accuracy, format, and hybrid-thinking objectives. Specifically:

\begin{itemize}
\item \textbf{Accuracy Reward} After removing the \texttt{<think>} tags, the output is compared against the ground truth. A match yields a reward of 1; otherwise 0.
\item \textbf{Format Reward} we assign a score of \textbf{1} if the output strictly follows the structural requirements by enclosing the reasoning within \texttt{<think></think>} tags, and \textbf{0} otherwise.
\item \textbf{Hybrid-Thinking Reward} To balance inference cost and accuracy, we introduce a discriminative reward that determines whether reasoning is necessary. When the input comes from the simple seed bank, omitting reasoning (i.e., directly producing the conclusion with empty \texttt{<think>} tags) yields a reward of 1, while unnecessarily including reasoning yields 0.
\end{itemize}

Finally, we combine the three rewards into the overall score:

\begin{equation}
R = 0.8 \cdot R_{\text{acc}} + 0.1 \cdot R_{\text{format}} + 0.1 \cdot R_{\text{hyb}},
\end{equation}

\begin{figure}[t]
\begin{tcolorbox}[
  colback=white, 
  colframe=black, 
  colupper=gray!70, 
  fontupper=\footnotesize, 
  coltitle=white, 
  coltext=black, 
  boxrule=0.5mm, 
  title=Simple Question,
  left=1mm, 
  right=1mm 
]
Is this image real or fake?\\
Does the photo look authentic?\\
Is this object captured in reality?\\
Was this photo taken with a real camera?\\
Is the clip authentic or generated?\\
Does the picture look natural or fake?\\
Is this painting generated or authentic?\\
Is this image authentic or fake?\\
Was this generated by AI or not?\\
Is the image real?\\
Is the image real or fake? Please answer me directly.
\end{tcolorbox}
\caption{Simple Question for Seed Question Set.}
\label{fig:simple}
\end{figure}
\begin{figure}[t]
\begin{tcolorbox}[
  colback=white, 
  colframe=black, 
  colupper=gray!70, 
  fontupper=\footnotesize, 
  coltitle=white, 
  coltext=black, 
  boxrule=0.5mm, 
  title=Hard Question,
  left=1mm, 
  right=1mm 
]
Explain why this image is real or fake.\\
Describe the visual artifacts that make this frame suspicious.\\
Justify whether this scene is realistic or synthetic.\\
What clues indicate that this screenshot is not authentic?\\
Provide reasoning for why the image looks artificial.\\
What makes this scenery photo appear fake or real?\\
Explain what features indicate that the shadows are unrealistic.\\
Which inconsistencies lead you to believe this image is fake?\\
What evidence supports that this video or image was generated by AI?\\
Why does the texture suggest this object is not real?\\
Is the image real or fake? Explain the reason.
\end{tcolorbox}
\caption{Hard Question for Seed Question Set.}
\label{fig:hard}
\end{figure}

\section{Data Formulation}

Current image generators are predominantly based on GANs and diffusion models. In the literature, existing studies on detecting AI-generated images are mostly limited to training on images or prompts from a single generative model. Overall, this problem setting faces two major challenges: (i) the training data is overly simplistic, as binary classification data obtained from existing images often fails to generalize to newly released generative models; and (ii) restricting model training to either binary classification data or reasoning data alone limits the model’s capacity for adaptive reasoning. To address these issues, we design a dual-mode dataset and leverage diverse data sources to support reinforcement learning research.

As shown in Table~\ref{table:chameleon}, demonstrating that the base models themself lack inherent capability in recognizing generated images. Therefore, in the SFT stage, we adopt the GenImage~\cite{genimage} dataset as the source of non-reasoning data without filtering, while for reasoning data, we prioritize interpretable datasets such as FakeClue~\cite{wen2025spot}. To avoid data leakage, we exclude Chameleon-related data from the FakeClue training set. Ultimately, the SFT stage consists of 1.834M non-reasoning samples and 94.188K reasoning samples.

For reinforcement learning, we build on the previous checkpoint and apply reject sampling~\cite{liu2024statisticalrejectionsamplingimproves}. Specifically, each sample in the FakeClue training set is evaluated five times, and samples correctly predicted in all runs are discarded. This results in 15.712K samples used in the RL stage. The question set (User Input) can be seen in Figure \ref{fig:simple} and \ref{fig:hard}.

Throughout training, instances are formatted according to their designated mode: (1) For reasoning data, the response must include a complete reasoning process, formatted as:\verb|<think>|\textbackslash nReasoning Steps\textbackslash n\verb|</think>|answer; (2)For non-reasoning data, the think tag is preserved in the following format: \verb|<think>|\verb|</think>|answer.


\begin{figure}[t]
\begin{tcolorbox}[
  colback=white, 
  colframe=black, 
  colupper=gray!70, 
  fontupper=\footnotesize, 
  coltitle=white, 
  coltext=black, 
  boxrule=0.5mm, 
  title=System Prompt,
  left=1mm, 
  right=1mm 
]
You are a helpful assistant for AI-generated image detection. Inspect the image and decide if it is real or fake.\\
Reasoning mode:\\
- If the image shows **obvious AI-generation traces** and is easy to detect, give no think steps.\\
- Otherwise, provide **careful, step-by-step** reasoning. If the image is easy to detect, output no think steps: \verb|<think>\n\n</think>|\\
\\
real or fake. If the user requests explain the think or the image is hard to detect, output the think steps: \\
\verb|<think>|\\
\verb|[Your reasoning here]| \\
\verb|</think>\n\nreal or fake|.
\end{tcolorbox}
\caption{System Prompt for Experiment and Training.}
\label{fig:system}
\end{figure}

\section{Experiment}

\subsection{Baselines}

\textbf{Baselines} We used Qwen2.5-VL-7B~\cite{wu2025qwenimagetechnicalreport} as the base model for training. We choose two open-source models, Qwen2.5-VL-7B and InternVL3-8B \cite{zhu2025internvl3exploringadvancedtraining}. The result of GPT-4o \cite{openai2024gpt4technicalreport} is from FakeClue \cite{wen2025spot}.

\textbf{Benchmarks} We evaluated models on the FakeClue test set~\cite{wen2025spot}, which is designed to assess generative image detection capability.

\textbf{Training and Evaluation Protocol} Our model was trained using the AdamW optimizer~\cite{adamw_Loshchilov2017FixingWD}. During the SFT stage, training was conducted for one epoch with a batch size of 1 and an initial learning rate of 1e-6, ensuring stable convergence on large-scale multimodal data. The RL stage was trained for three epochs with a batch size of 32, using the same learning rate (1e-6). The final $\method_{rl}$ model was obtained by applying the RL process to the last SFT checkpoint. Training required approximately 20 hours for SFT and 40 hours for RL on a system equipped with 32 A100 GPUs. The system prompt used for training and evaluation is shown in Figure~\ref{fig:system}.

\begin{table}[htb]
\centering
\resizebox{\linewidth}{!}{
\begin{tabular}{ccccccc}
\toprule
\multirow{2}{*}{Method} & \multicolumn{2}{c}{Real} & \multicolumn{2}{c}{Fake} & \multicolumn{2}{c}{Overall}\\ 
\cmidrule(r){2-3} \cmidrule(r){4-5} \cmidrule(r){6-7}
 & Acc & F1 & Acc & F1 & Acc & F1 \\
 \midrule
 GPT-4o \cite{openai2024gpt4technicalreport} & - & - & - & - & 47.40 & 42.00 \\
Qwen2.5-VL-7B \cite{wu2025qwenimagetechnicalreport} & 81.91 & 52.44 & 9.05 & 16.5 & 35.40 & 29.50 \\
InternVL3-8B \cite{zhu2025internvl3exploringadvancedtraining} & 69.14 & 57.04 & 48.31 & 60.49 & 55.84 & 59.24 \\
\midrule
Fake-HR1$_{sft\_noreasoning}$ & \textbf{100.00} & 72.45 & 56.92 & 72.55 & 72.50 & 72.51 \\
Fake-HR1$_{sft\_reasoning}$ & 99.83 & 72.94 & 57.11 & 72.70 & 72.56 & 72.79 \\
Fake-HR1$_{hrl\_noreasoning}$ & 91.65 & 83.52 & \textbf{84.24} & \textbf{89.16} & \textbf{86.92} & \textbf{87.12} \\
Fake-HR1$_{hrl\_reasoning}$ & 94.41 & \textbf{84.15} & 81.17 & 88.16 & 85.96 & 86.71 \\
\bottomrule
\end{tabular}
}
\caption{Comparison with other detection methods or VLMs on the FakeClue test dataset.}
\label{table:chameleon}
\end{table}

\subsection{Main Results}

Table~\ref{table:chameleon} reports the performance of all models on the fakeclue test dataset. At the 7B scale, $\method_{rl}$ consistently outperformed all baselines and its SFT counterpart. This improvement indicates that HRL fine-tuning effectively mitigates the bias toward Real classification and enables the model to generalize more reliably across diverse generative distributions. Our experiments are divided into reasoning and non-reasoning settings: $\method_{noreasoning}$ denotes training with simple seed questions, while $\method_{reasoning}$ uses hard seed questions; $\method_{sft}$ and $\method_{hrl}$ follows the same division. Notably, for the same model with reasoning switched on or off, the reasoning variants achieve slightly lower scores compared to their non-reasoning counterparts.

\begin{figure}[htbp]
    \centering
    \includegraphics[width=0.95\columnwidth]{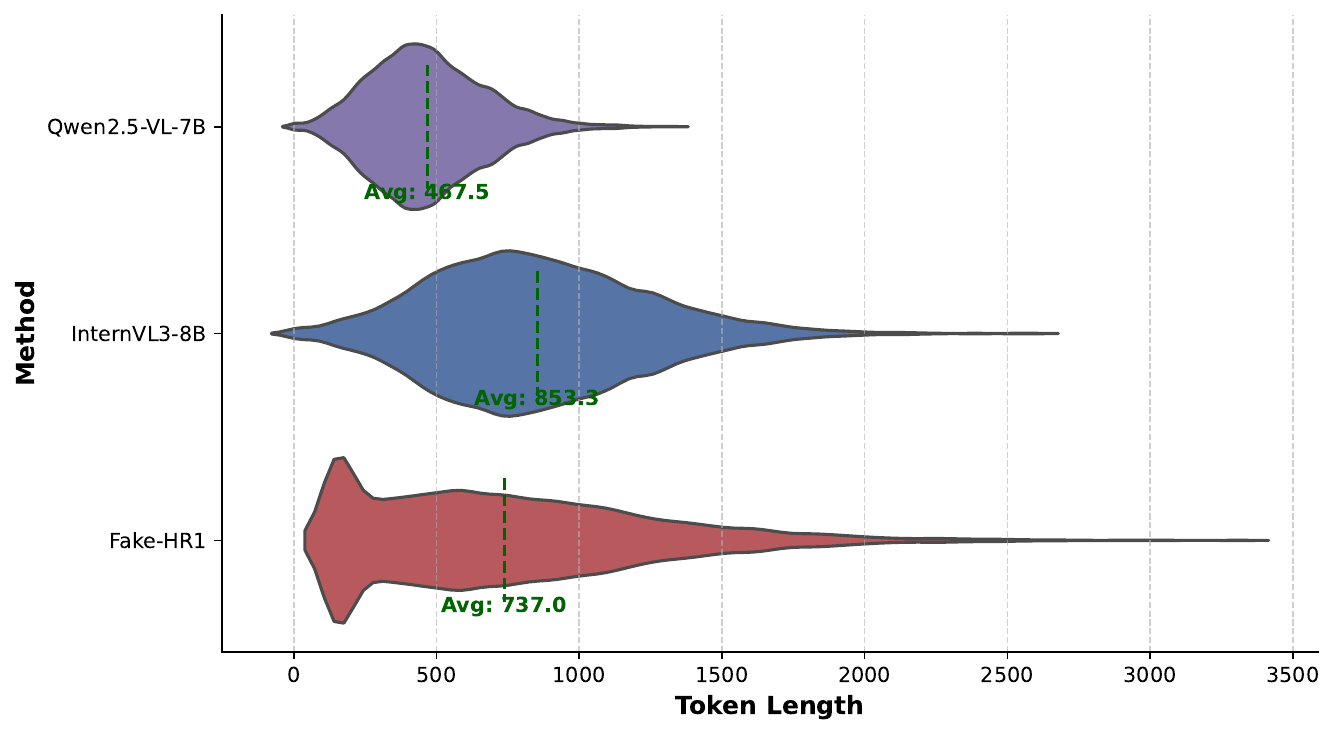}
    \caption{Distribution of output token lengths. \method~is $\method_{reasoning}$.}
    \label{fig:token_length}
\end{figure}

Figure~\ref{fig:token_length} illustrates the distribution of output token lengths across different models. Since many outputs contained repeated segments that inflated token counts, we excluded such samples from the manual statistics for \method. \method~reduces the average token length by aligning the model with more concise human annotations, yet still exhibits considerable variance. $\method_{hrl\_noreasoning}$ \textbf{consistently produces an output length of 23 tokens}, maintaining high classification accuracy while generating concise responses. This demonstrates the key advantage of \method: it can dynamically decide whether reasoning is necessary, thereby saving both output tokens and inference time.

\begin{figure}[!hp]
    \centering
    \includegraphics[width=0.95\columnwidth]{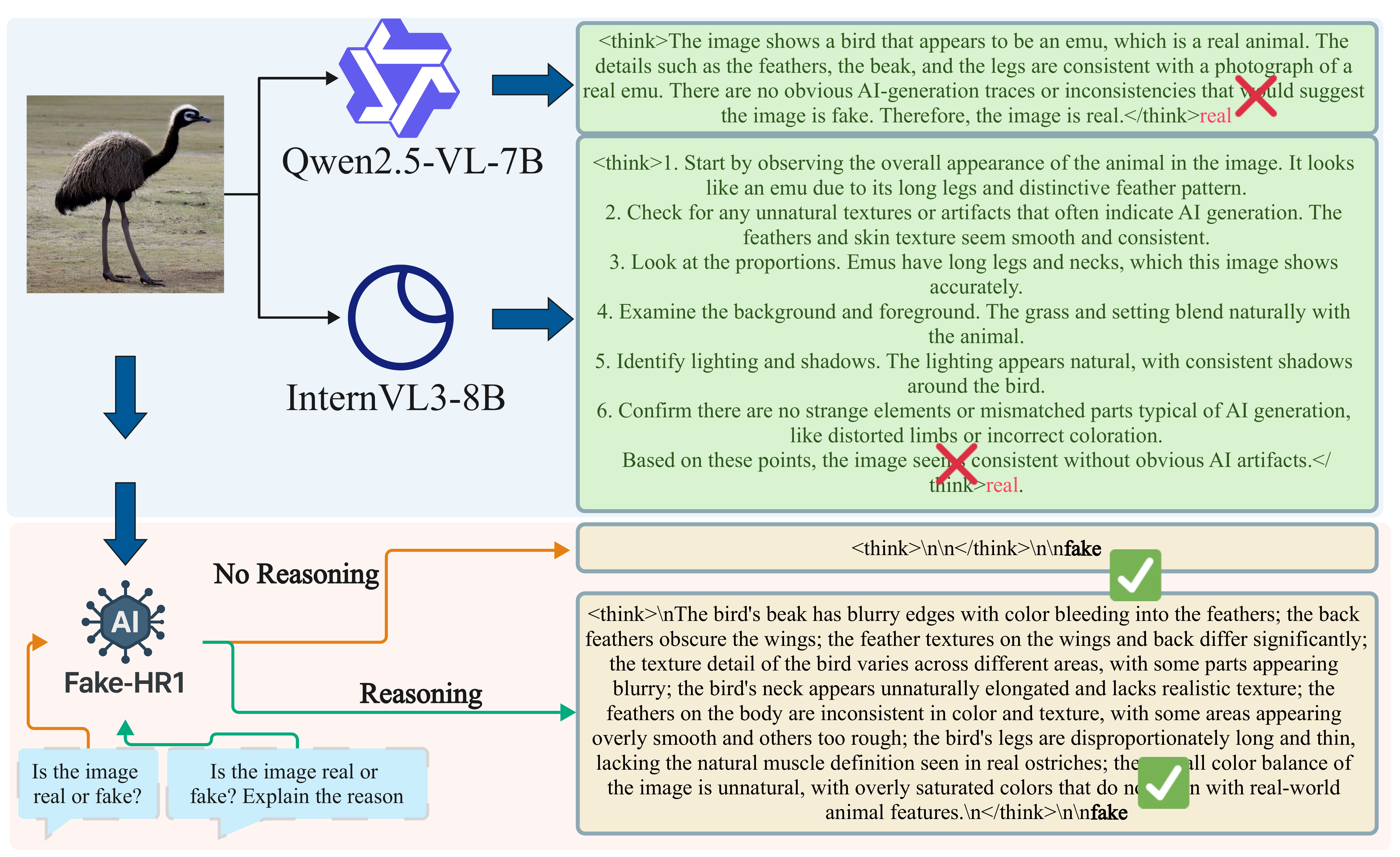}
    \caption{Case Study for \method.}
    \label{fig:case}
\end{figure}

\subsection{Case Study}

As shown in Figure~\ref{fig:case}, \method~can automatically switch between reasoning modes depending on the difficulty of the question. 
In contrast, baseline models often generate overly lengthy explanations that inflate output cost without improving accuracy. These examples demonstrate the superiority of our method in generative detection: it achieves reliable performance across varying levels of task difficulty while substantially reducing inference cost. Prior work such as Loki~\cite{ye2024loki} shows that both open-source and proprietary LLMs lack intrinsic AIGC-detection capability; consequently, their predictions are often unreliable without specialized training. This reinforces the necessity of dedicated AIGC-detection models. In contrast, \method~produces explanations that remain both concise and fine-grained, effectively balancing interpretability with output efficiency while maintaining strong discriminative performance.

\section{Conclusion}

In this work, we aimed to develop a MLLM capable of effectively balancing reasoning ability and synthetic image detection performance. To this end, we proposed a two-stage training framework consisting of SFT and HGRPO. Experimental results demonstrate that this framework substantially improves detection performance while simultaneously enhancing hybrid reasoning. Specifically, it reduces unnecessary reasoning on simple queries-an inefficiency often observed in LRMs and mitigates the insufficient reasoning capacity commonly found in traditional MLLM. By addressing the critical bottleneck of inference inefficiency in real-world AIGC detection, \method~further strengthens practical applicability in deployment-oriented scenarios.

For future work, we plan to further explore the varying levels of complexity in synthetic text \cite{hu2023radar}, images and videos \cite{aigvdet,zhang2025ivyfakeunifiedexplainableframework,effort}. Beyond the difficulty of queries themselves, hybrid reasoning could adaptively determine whether reasoning is required based on the inherent difficulty of the synthetic image detection task. 

Moreover, while the current approach relies on a fixed seed question set, we plan to adopt an online, large-model–generated seed dataset to enhance the diversity and adaptability of generated queries.

\section{RELATION TO PRIOR WORK}
\label{sec:prior}

Compared with prior works, our framework introduces several key innovations. First, unlike FakeVLM~\cite{wen2025spot}, FakeShield~\cite{xu2024fakeshield}, IvyFake~\cite{zhang2025ivyfakeunifiedexplainableframework} and UniShield~\cite{unishield}, which rely on excessively long reasoning chains, our approach does not require such overextended reasoning when dealing with images that exhibit clear generative artifacts. Instead, it can autonomously decide whether reasoning is necessary based on the difficulty of the task and the characteristics of the AI-generated image. Second, inspired by AIDE~\cite{yan2025sanitycheckaigeneratedimage}, SIDA~\cite{kim2025sidasyntheticimagedriven} and DiffusionFake~\cite{sun2024diffusionfake}, which are limited to binary classification, our method further produces interpretable reasoning chains, thereby making the classification results more trustworthy. Finally, unlike existing hybrid reasoning approaches such as Qwen3 \cite{yang2025qwen3technicalreport}, where hybrid reasoning is constrained by fixed prompt tokens (\texttt{\textbackslash think}, \texttt{\textbackslash nothink}) and the model cannot adaptively determine the necessity of reasoning, our framework—drawing inspiration from \cite{jiang2025thinkneedlargehybridreasoning}—introduces diverse queries during data construction and integrates a hybrid reasoning reward in the RL stage. This enables the VLM to autonomously decide whether reasoning is required.

\section{Acknowledgments}

This research is supported by the Young Scientists Fund of the National Natural Science Foundation of China (Grant No.72304215) and the Ant Group Research Intern Program.

\vfill\pagebreak

\bibliographystyle{IEEEbib}
\bibliography{Template}

\end{document}